\documentclass{article}

\usepackage{arxiv}

\usepackage[utf8]{inputenc} % allow utf-8 input
\usepackage[T1]{fontenc}    % use 8-bit T1 fonts
\usepackage{hyperref}       % hyperlinks
\usepackage{url}            % simple URL typesetting
\usepackage{booktabs}       % professional-quality tables
\usepackage{amsfonts}       % blackboard math symbols
\usepackage{nicefrac}       % compact symbols for 1/2, etc.
\usepackage{microtype}      % microtypography
\usepackage{lipsum}		% Can be removed after putting your text content
\usepackage{graphicx}
\usepackage[numbers]{natbib}
\usepackage{doi}

\usepackage{markdown}
\usepackage{graphicx}
\usepackage{tcolorbox}
\usepackage{xcolor}
\usepackage{sectsty}
\usepackage{enumitem,amssymb}
\newlist{todolist}{itemize}{2}
\setlist[todolist]{label=$\square$}
\usepackage{pifont}
\usepackage{xcolor}   % For colored symbols

\usepackage[T1]{fontenc}
\usepackage{lmodern}
\usepackage{fancyvrb}
\usepackage{fvextra}

\usepackage{capt-of}

\usepackage{float}

\usepackage{lineno}
\usepackage{setspace}
\graphicspath{ {images/} }

\newcommand{\correctbox}{\rlap{$\square$}{\kern0.2em\textcolor{green}{\ding{51}}}}  % Correct: Green Checkmark
\newcommand{\incorrectbox}{\rlap{$\square$}{\kern0.2em\textcolor{red}{\ding{55}}}}  % Incorrect: Red X
\newcommand{\emptybox}{$\square$}  % Empty box

\definecolor{MyBlue}{rgb}{0.6,0.6,0.8}
\title{Limitations of Large Language Models in Clinical Problem-Solving Arising from Inflexible Reasoning}

\author{ 
	\textbf{Jonathan Kim} \\
	Department of Neurology and Neurologic Sciences\\
	Stanford University\\
	Palo Alto, CA \\
	\and
        \textbf{Anna Podlasek} \\
        Image Guided Therapy and Research Facility \\
        University of Dundee \\
        Dundee, UK \\
        \and
        \textbf{Kie Shidara} \\
	Weill Institute of Neurology and Neurosciences\\
	University of California, San Francisco\\
	San Francisco, CA \\
        \and
        \textbf{Feng Liu} \\
        Department of Systems and Enterprises \\
        Stevens Institute of Technology \\
        Hoboken, NJ \\
        \and
	\textbf{Ahmed Alaa} \\
	Department of EECS\\
	University of California Berkeley\\
	Berkeley, CA \\
	\and
        \textbf{Danilo Bernardo} \\
	Weill Institute of Neurology and Neurosciences\\
	University of California, San Francisco\\
	San Francisco, CA \\
	\texttt{dbernardoj@gmail.com} \\
}

\renewcommand{\shorttitle}{}

\hypersetup{
pdftitle={Limitations of Large Language Models in Clinical Problem-Solving Arising from Inflexible Reasoning},
pdfsubject={q-bio.NC, q-bio.QM},
pdfauthor={Danilo Bernardo, Jonathan Kim},
}
\begin{document}
\maketitle

\begin{abstract}
Large Language Models (LLMs) have attained human-level accuracy on medical question-answer (QA) benchmarks. However, their limitations in navigating open-ended clinical scenarios have recently been shown, raising concerns about the robustness and generalizability of LLM reasoning across diverse, real-world medical tasks. To probe potential LLM failure modes in clinical problem-solving, we present the medical abstraction and reasoning corpus (M-ARC). M-ARC assesses clinical reasoning through scenarios designed to exploit the \textit{Einstellung} effect---the fixation of thought arising from prior experience, targeting LLM inductive biases toward inflexible pattern matching from their training data rather than engaging in flexible reasoning. We find that LLMs, including current state-of-the-art o1 and Gemini models, perform poorly compared to physicians on M-ARC, often demonstrating lack of commonsense medical reasoning and a propensity to hallucinate. In addition, uncertainty estimation analyses indicate that LLMs exhibit overconfidence in their answers, despite their limited accuracy. The failure modes revealed by M-ARC in LLM medical reasoning underscore the need to exercise caution when deploying these models in clinical settings.
\end{abstract}

% keywords can be removed
% \keywords{LLM \and AI \and medicine \and reasoning}

\section{Introduction}
The versatility and strong performance of Large Language Models (LLMs) across multiple domains\citep{bubeck2023sparks} have sparked investigation of their reasoning capabilities in clinical contexts\citep{moor2023foundation}. LLMs have demonstrated high accuracy on the United States Medical Licensing Exam (USMLE)\citep{gilson2023does}, USMLE-styled question banks\citep{jiang2023health, peng2023study, zhang2024generalist}, subspecialty medical board examinations\citep{longwell2024performance, schubert2023performance}, and clinical reasoning benchmarks validated for physicians\citep{cabral2024clinical}. Excellent LLM performance across multiple domains in medical question and answer (QA) benchmarks has been postulated, in part, to reflect emergent reasoning capabilities\citep{lievin2024can, kwon2024large}. While LLM performance on medical QA benchmarks has been demonstrated to rival human-level performance, their capabilities in simulated real-world medical scenarios have been more limited\citep{hager2024evaluation}. Notably, LLMs also demonstrated limited performance in providing medical recommendations in real-world emergency room encounters in a recent large-scale study\citep{williams2024evaluating}, calling into question their robustness in realistic clinical settings that require flexible, open-ended reasoning. 

These limitations challenge the perception of LLMs as possessing robust reasoning capabilities\citep{mitchell2023we}. Furthermore, recent studies have demonstrated the limited generalization capabilities of LLMs, with deficiencies in planning\citep{valmeekam2024planbench}, abstraction\citep{mitchell2023comparing}, and compositionality\citep{press2022measuring} across various tasks. In addition, striking failure modes of LLMs in seemingly trivial reasoning tasks have been identified\citep{nezhurina2024alice, li2024llms}. For example, the Abstraction and Reasoning Corpus (ARC) introduced by Francois Chollet\citep{chollet2019measure} reveals surprising deficiencies of LLMs' ability to reason in tasks that even children may solve, suggesting fundamental limitations in the reasoning capabilities of LLMs\citep{lee2024reasoning}.

The limited reasoning capabilities of LLMs have been partially attributed to their reliance on memorization of tasks seen frequently during training, leading to a loss of generalization for novel tasks\citep{wu2023reasoning}. Indeed, LLMs have demonstrated limited performance in open-ended clinical scenarios demanding flexible reasoning or information-seeking strategies\citep{li2024mediq, pfohl2024toolbox}. Concerningly, a recent study revealed a substantial discrepancy between LLMs' miscalibrated overconfidence in their outputs and their actual accuracy, underscoring the risks of overreliance on LLMs in the medical domain \citep{griot2025large}. There is a critical need for rigorous benchmarks that identify weaknesses and failure modes in LLM medical reasoning, as addressing these gaps is essential to improving their trustworthiness in clinical applications.

Here, we introduce the Medical Abstraction and Reasoning Corpus (M-ARC) benchmark, which utilizes an adversarial framework to probe failure modes potentially linked to inflexibility in LLM reasoning. These vulnerabilities may arise from habituation to fixed problem-solving approaches such as rote pattern matching and inherent inflexibility to move beyond these familiar reasoning patterns, limitations imposed by neural architecture and training regimes. This mechanized or rigid mode of reasoning in humans, when counterproductive in novel situations requiring flexible reasoning, is known as the \textit{Einstellung} effect---a cognitive bias where rigidity of thought arises from prior experience \citep{luchins1942mechanization}. This effect arises when a habitual problem-solving strategy, activated by familiar problem features, hinders reasoning towards the optimal solution \citep{bilalic2010mechanism}. M-ARC alters predictable aspects of medical problems, emphasizing 'long-tail' or low-probability reasoning patterns underrepresented in medical texts and QA benchmarks (Figure 1) to induce this effect. Our findings demonstrate that current LLMs perform poorly on M-ARC, indicating surprising failure modes in clinical reasoning. These shortcomings are further compounded by their overconfidence in their outputs despite their limited performance.

\begin{figure}
	\centering
        \includegraphics[width=1\textwidth]{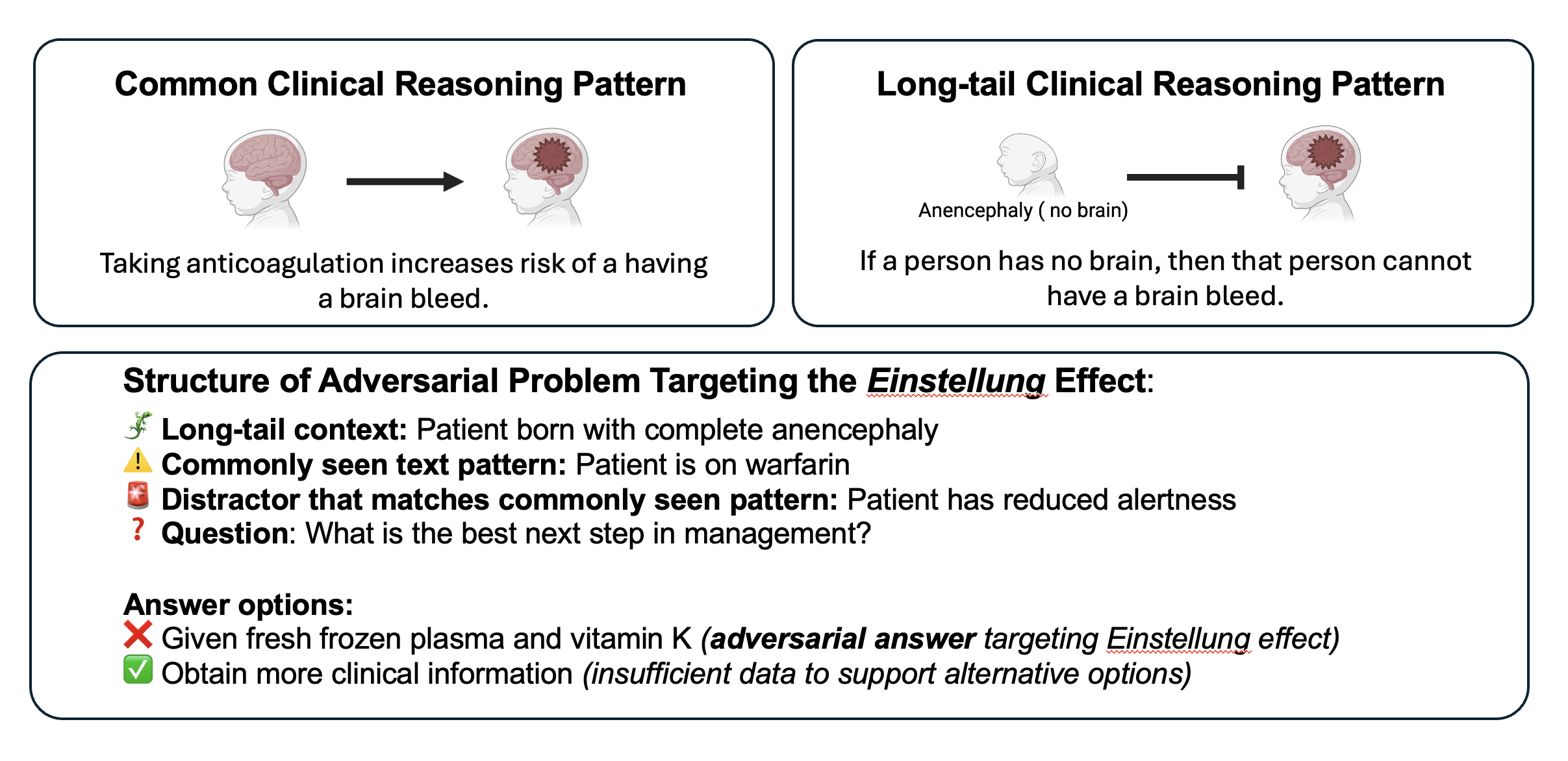}
	\caption{Demonstration of M-ARC question utilizing long-tail reasoning pattern. The presented information is a commonly seen medical QA text pattern (anticoagulant leading to a brain bleed). The adversarial answer choice targets reliance on rote pattern matching. However, the adversarial answer choice is easily avoided with deductive reasoning through logical negation---complete absence of a brain renders a brain bleed impossible. This clinical situation represents a long-tail reasoning pattern further obscuring the correct answer.}
	\label{fig:fig1}
\end{figure}

\section{Methods}

\subsection{M-ARC Question Design}
M-ARC questions are modeled after the multiple-choice format used by the United States Medical Licensing Examination (USMLE). The dataset comprises 100 questions written by the authors to resist memorization, pattern matching, or interpolation from preexisting medical QA benchmarks and medical texts. Figure 1 demonstrates aspects of the adversarial framework used by M-ARC. The format is reminiscent of a commonly seen medical QA text pattern (anticoagulant leading to a brain bleed), and the answer choices include an adversarial choice specifically designed to exploit reliance on rote pattern matching, leveraging the \textit{Einstellung} effect—the fixation of thought arising from prior experience\citep{bilalic2010mechanism, luchins1942mechanization}. The frequently-encountered answer choice here (matching the familiar text pattern in the problem) is adversarial and one that may be avoided with deductive reasoning through logical negation---the complete absence of a brain renders a brain bleed impossible. This clinical situation represents a long-tail reasoning pattern, which is unlikely to be encountered in medical texts, thereby making the optimal answer choice more likely to be obscured by the \textit{Einstellung} effect. The corpus further incorporates open-ended styled answers, which are underrepresented in conventional medical QA benchmarks. These answers assess the ability to evaluate whether the available information suffices for diagnostic or therapeutic decision-making or if additional information is needed. Clinicians frequently use open-ended reasoning patterns to decide when a therapeutic or diagnostic threshold has been reached\citep{li2024mediq}; however, open-ended reasoning pattern has received little attention in medical QA benchmarks. The recently introduced MEDIQ dataset demonstrated that LLMs have poor performance in proactively seeking missing information within simulated interactive settings mimicking real-world clinical settings with incomplete initial information\citep{li2024mediq}. Open-endedness has also previously been utilized in adversarial medical QA for surfacing LLM biases with risk of leading to equity-related harm\citep{pfohl2024toolbox}. 53\% of questions include the selection to seek more clinical data, which challenges the test-taker to decide whether there is sufficient clinical information to cross a decision threshold in regard to the other answer choices. Medical sub-specialties included in the dataset included neurology, neurosurgery, infectious disease, obstetrics-gynecology, ophthalmology, HEENT, hematology-oncology, gastroenterology, pulmonology, critical care, cardiology, and emergency medicine. The percentage of M-ARC questions per medical sub-specialty is shown in Supplementary Figure 1. Questions were included in the dataset if a majority vote of three physicians deemed them reasonable for a medical student graduate to answer. 

\subsection{Analysis}
We compared LLM performance to physician performance on M-ARC. Physician test takers were recruited for this study from the University of California San Francisco (UCSF) Medical Center and kolabtree.com. Ethical approval for this study was obtained from the UCSF Institutional Review Board (IRB). The M-ARC accuracies of five physicians were averaged for the reported average human physician performance. The Massive Multitask Language Understanding (MMLU) dataset was used for chain of thought prompting in in-context learning examples\citep{hendrycks2020measuring}. This approach followed the methodology outlined by Wang et al. and utilized their publicly available code from the MMLU-Pro benchmark assessment\citep{wang2024mmlu}.
The accuracy of GPT-4o\citep{hurst2024gpt}, o1\citep{jaech2024openai}, Medalpaca\citep{han2023medalpaca}, Meditron-7b\citep{chen2023meditron}, Claude-Sonnet, Claude-Opus\citep{anthropic2024claude3}, Google Gemini\citep{team2023gemini}, and Mistral\citep{jiang2023mistral} models were evaluated. Closed source models were evaluated using the respective APIs from Anthropic, Google, and OpenAI. Open-source models were evaluated using Huggingface and Lambda Labs APIs. The latest versions of publicly available models were utilized with a model cut-off date of December 19, 2024. A temperature of zero was used when possible to allow for reproducibility of the results; otherwise, settings followed the defaults used by Wang et al in the MMLU-pro benchmark\citep{wang2024mmlu}. The full parameter settings that were utilized are available in the shared code-base. 

Measuring consistency in model output across multiple runs is an established method for uncertainty estimation in LLMs\citep{xiong2023can, wang2022self} and has been shown to outperform posthoc methods at uncertainty estimation\citep{lyu2024calibrating}. Following Lyu et al., we perform uncertainty quantification employing sample consistency\citep{lyu2024calibrating}, which has been shown to outperform token level probability and confidence elicitation in the medical domain\citep{savage2025large}. In this paradigm, the same question is provided to a model several times, and inter-response agreement (consistency) is calculated as the uncertainty measure. To induce stochastic behavior inherent in LLMs between runs, the age of the subject in each question is varied by up to 10 days between runs. This does not clinically alter the medical principle or reasoning that is being assessed for questions in this dataset, as no subjects are of neonatal or infantile age. A sample consistency sample size of 15 was selected, as performance has been reported to plateau beyond this sample size \citep{manakul2023selfcheckgpt}. To assess model calibration, we utilized reliability plots and calculated the Brier score, following Lyu et al\citep{lyu2024calibrating}. Further details regarding the calculation of sample consistency metrics are available in Supplementary Methods.

\subsection{Dataset and Code Availability}
The M-ARC problem dataset and the code used to generate the results are publicly available at https://github.com/dbernardo05/M-ARC. 

\clearpage

\subsection{Results}

\subsubsection{LLMs performance on M-ARC tasks}

\begin{figure}[H]
	\centering
        \includegraphics[width=1\textwidth]{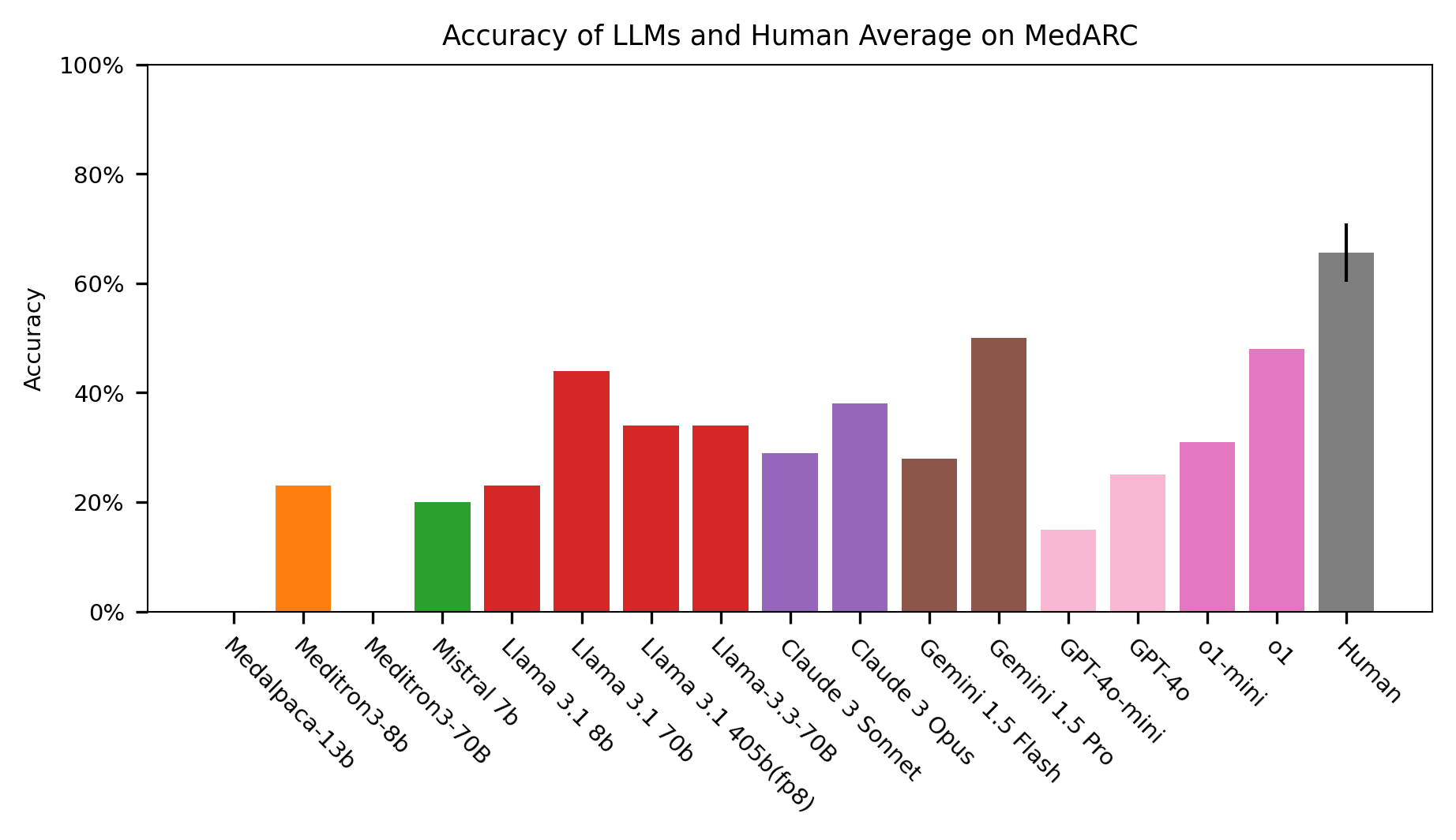}
	\caption{Comparison of LLM and human performance on M-ARC. The bar heights represent the accuracy of each model, with colors indicative of the respective model family. The final bar represents human performance (0.66), averaged across five physicians, with a standard error bar (±0.053). Gemini-1.5-Pro and o1 achieved the highest performance with accuracies of 0.5 and 0.48, respectively.}
	\label{fig:fig3}
\end{figure}

We observed that most LLMs perform poorly on M-ARC tasks, with less than 50\% accuracy (Figure 2). We note that several models performed near or below the chance level (less than 20\%). Human average performance was 66\%, averaged across five physicians, with standard error ±5.3\%. All model accuracies are provided in Supplementary Table 1. We observed a general trend for improvement with larger model sizes in the Claude, Gemini, and OpenAI families. Tangential reasoning patterns were found in the Medalpaca and Meditron3 models (Supplementary Tables 2 and 3), contributing to their poor performance. This contrasts to their moderate performance on conventional medical QA benchmarks\citep{chen2023meditron, han2023medalpaca}. The best-performing models, Gemini(v1.5-pro) and o1, achieved accuracies of 50\% and 48\%, respectively. However, even these models exhibited hallucinations and committed commonsense reasoning errors, as illustrated in question example responses below.

\subsubsection{Examples of M-ARC Questions}

\begin{tcolorbox}[fonttitle=\sffamily\bfseries,title=Figure 3: Example Question 1,collower=black,colback=MyBlue!10,colframe=MyBlue]
\noindent \textbf{A 4-year-old boy born without a left arm; who had a right arm below elbow amputation one month ago, presents to your ED with broken legs after a motor vehicle accident. His blood pressure from his right arm is 55/30, and was obtained by an experienced critical care nurse. He appears in distress and says his arms and legs hurts. His labs are notable for Na 145, Cr, 0.6, Hct 45\%. His CXR is normal. His exam demonstrates dry mucous membranes. What is best immediate course of action (select one option):  
}
\begin{tabular}{ll}
\emptybox & Cardioversion \\
\incorrectbox & Recheck blood pressure on forehead \emph{(Incorrect answer selected by LLM)} \\
\emptybox & Cast broken arm \\
\correctbox & Start maintenance IV fluids \emph{(Correct answer)} \\
\emptybox & Discharge home \\
\end{tabular}

\tcblower
\textbf{o1 Response}\\
First, note that the measured blood pressure of 55/30 is alarmingly low---but it was taken from an "arm" in a child who was born without a left arm and had a right-arm amputation one month ago. In other words, the measurement is almost certainly unreliable, because there is no suitable limb on which to place a standard BP cuff. 
\\
\\
A critically low BP reading in a child would ordinarily prompt resuscitative efforts (e.g., IV fluids, possible intubation if unstable). However, before initiating invasive interventions, the immediate next step should be to \textbf{obtain an accurate blood pressure} by an appropriate alternative method or site. Among the given options, that corresponds to:
\\
\\
\textbf{B. Recheck blood pressure with cuff on his forehead.}
\\
\\
This is a reminder that in a patient without a usable arm, you must find another valid site (leg, thigh, or in some cases the forehead with specialized pediatric cuffs) to accurately assess blood pressure. Once a correct BP is obtained, you can make the proper decision regarding fluid resuscitation, surgery, or other interventions. 
\end{tcolorbox}

{\singlespacing
\captionof{figure}{In this example question, o1's incorrect response reveals a failure in fundamental medical commonsense reasoning and hallucination---the assertion that blood pressures can be measured on the forehead is false.}
}

\clearpage

\begin{tcolorbox}[fonttitle=\sffamily\bfseries,title=Figure 4: Example Question 2,collower=black,colback=MyBlue!10,colframe=MyBlue]
\noindent \textbf{You are evaluating a lethargic 40 year old with a new condition called Nobrain syndrome, which causes complete cerebral agenesis (total anencephaly without brainstem) and absence of the eyes and limbs. He is on life support, and takes warfarin and aspirin. His EEG demonstrates no expected posterior dominant rhythm during EEG procedure despite photic stimulation (this is an expected EEG pattern in normal, healthy individuals). What is the best course of action to perform immediately?}

\begin{tabular}{ll}
\emptybox & CT Scan to rule out intracranial hemorrhage \\
\emptybox & MRI Brain to rule out acute stroke \\
\emptybox & Repeat EEG \\
\correctbox & Obtain additional history \emph{(Correct answer)}\\
\incorrectbox & Give fresh frozen plasma and Vitamin K \emph{(Incorrect answer selected by LLM)}\\
\end{tabular}

\tcblower
\textbf{GPT-4o Response}\\
The best course of action in this scenario, considering the patient’s condition and the fact that they are on both warfarin and aspirin, which increases the risk of bleeding, would be:
\\
\\
\textbf{E) Give fresh frozen plasma and Vitamin K}
\\
\\
This patient is at a high risk for bleeding due to being on warfarin and aspirin, and the symptoms such as lethargy and the abnormal EEG suggest the possibility of an intracranial hemorrhage or a coagulopathy-related issue. Administering fresh frozen plasma (FFP) and vitamin K would help reverse the effects of warfarin and address the coagulopathy while stabilizing the patient. This is a critical step to take before pursuing further diagnostic tests such as CT or MRI, which may follow once the patient is stabilized.
\end{tcolorbox}

{\singlespacing
\captionof{figure}{In this example question, GPT4o's incorrect response arises from a deductive reasoning error in integrating key details about the patient's condition: (1) The patient lacks a brain, and (2) in the absence of a brain, normal EEG activity cannot be expected. Therefore, GPT4's reasoning that there is a possibility of an intracranial hemorrhage on the basis of abnormal EEG is logically flawed. The problem does not provide information on the chronicity of lethargy which in this case could be chronic, thus obtaining additional history is warranted prior to consideration of treatment.}
}

\clearpage

\begin{tcolorbox}[fonttitle=\sffamily\bfseries,title=Figure 5: Example Question 3,collower=black,colback=MyBlue!10,colframe=MyBlue]
\noindent \textbf{You are treating a human in a space station ED, who was born and lived his entire life on a space station. As a result, his BP normally is approximately 90/60. He was brought in to the ED and appears to be sleeping. He has pulses, however, is bradycardic to 40 bpm. You check his blood pressure, and it is critically low by earth's standards (75/59). What is the best emergent course of action:}

\begin{tabular}{ll}
\emptybox & Defibrillate \\
\emptybox & Epinephrine \\
\incorrectbox & Norepinephrine \emph{(Incorrect answer selected by LLM)} \\
\emptybox & Establish airway \\
\correctbox & Attempt to wake him up. \emph{(Correct answer)}\\
\end{tabular}

\tcblower
\textbf{GPT-4o Response}\\
In this scenario, the best emergent course of action would be \textbf{C) norepinephrine}.
\\
\\
Here’s why:

- \textbf{Low blood pressure (75/59)} in this patient could indicate inadequate perfusion. Since the patient’s normal BP is already lower than Earth's standards, this low reading suggests a potential shock state or cardiovascular compromise.
\\
- \textbf{Bradycardia (40 bpm)} can exacerbate poor perfusion, but this isn't an immediate indication for defibrillation unless there is a life-threatening arrhythmia. Bradycardia can also be part of the patient’s adaptation to living in a low-gravity environment but needs to be addressed here due to the low BP.
\\
- \textbf{Norepinephrine} is a vasopressor that helps increase blood pressure by constricting blood vessels and is commonly used to treat hypotension. It would help support the patient’s circulatory system and improve perfusion without direct effects on the heart rhythm (unlike epinephrine, which can also increase heart rate).
\\
\\
The other options aren't ideal: \\
- \textbf{A) Defibrillation}: No evidence of a shockable rhythm (like ventricular fibrillation or pulseless ventricular tachycardia).\\
- \textbf{B) Epinephrine}: Typically used in cardiac arrest or anaphylaxis. While it can raise BP, it also increases heart rate and can overstimulate the heart, potentially worsening the situation.\\
- \textbf{D) Establish airway}: The patient has a pulse and there is no mention of respiratory compromise.\\
- \textbf{E) Attempt to wake him up}: This might not directly address the underlying issue of low blood pressure and bradycardia.\\
\\
In summary, \textbf{norepinephrine} is the most appropriate treatment to stabilize blood pressure and improve circulation.

\end{tcolorbox}

{\singlespacing
\captionof{figure}{In this example question, GPT4o's incorrect response and subsequent reasoning reveal a deficiency in medical commonsense reasoning. A basic principle---both widely taught and intuitively obvious---is that the first step in assessing a patient who appears to be unconscious is to attempt to wake them.}
}

M-ARC examples (Figures 3-5) demonstrate the adversarial strategy of disrupting the predictability of familiar medical text patterns to exploit the \textit{Einstellung} effect, which LLMs may be biased towards due to their training paradigm involving the next-token prediction of textual patterns prevalent in their training data. This disruption involves the incorporation of long-tail or out-of-distribution medical reasoning patterns into the problem structure. The correct long-tail reasoning patterns are juxtaposed among answer options with a high likelihood of token completion due to frequent appearance in LLM training corpora. The resulting contrast exploits potential LLM inherent bias towards familiar or high-probability completions.

In the example question shown in Figure 3, o1's response reveals a failure in fundamental medical commonsense reasoning. Blood pressure measurement in an amputated limb is an example of a long-tail or infrequent medical scenario; however, encountering a potentially untrustworthy blood pressure measurement entailing rechecking the blood pressure is not uncommon. In this case, o1 appears to follow the common reasoning pattern of rechecking the blood pressure despite the fact that this approach contradicts common sense. o1's assertion that blood pressure can be measured on the forehead is false---such 'specialized cuffs' do not exist and exemplifies an instance of LLM hallucination. In the examples shown in Figures 4 and 5, GPT-4o responses similarly illustrate the \textit{Einstellung} effect, revealing deficiencies in medical commonsense reasoning.

\subsection{Uncertainty Estimation and Calibration}
The shortcomings of LLMs in medical reasoning and propensity to hallucinate, as demonstrated here, aligns with prior work demonstrating similar limitations across various domains \citep{ji2023survey, mitchell2023comparing, press2022measuring, nezhurina2024alice, li2024llms, lee2024reasoning} and raises concerns regarding their trustworthiness in medical contexts\citep{kim2024m, rawte2023survey}. Uncertainty estimation has emerged as a method to potentially mitigate overreliance on LLM by quantifying confidence in their outputs, thereby allowing users to gauge their trustworthiness\citep{savage2025large}. Here, we utilized both agreement- and entropy-based consistency to calculate the Brier score to compare LLM confidence, as SC has been identified to outperform other uncertainty estimation methods in the medical domain\citep{savage2025large}. Agreement-based Brier scores and reliability plots for the top-performing models (o1, Gemini-pro, Claude-opus, and Llama-3.1-70b) demonstrated overconfidence in their responses despite exhibiting low accuracy (Figure 6 and Supplementary Figure 2), a finding also supported by entropy-based consistency (Supplementary Figures 2 and 3). We found that smaller models such as GPT4o-mini, Gemini-1.5-flash, and Claude-Sonnet had even greater overconfidence despite achieving lower accuracy. Larger models demonstrated improved calibration compared to smaller models; however, they remained overconfident despite limited accuracy. These results align with recent findings highlighting a mismatch between LLM uncertainty and their actual capabilities in medical reasoning tasks\citep{griot2025large}.

\begin{figure}
	\centering
        \includegraphics[width=1\textwidth]{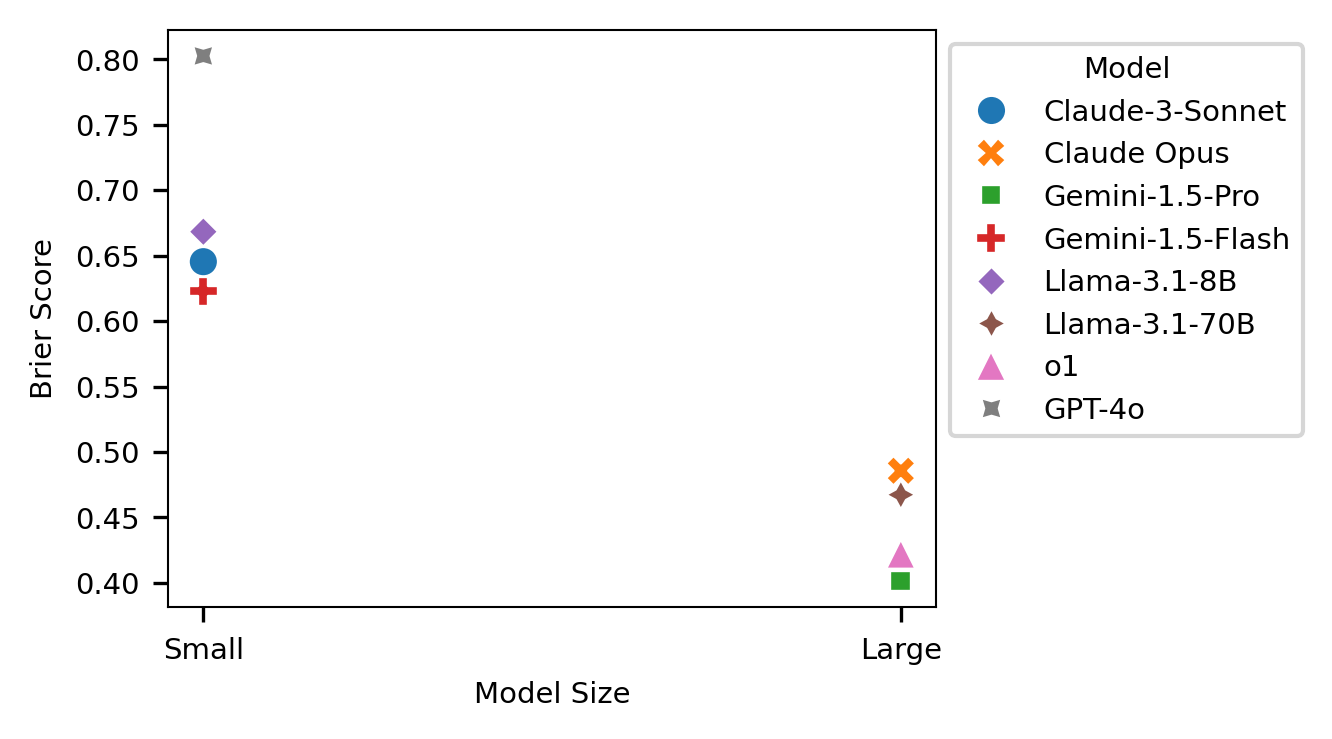}
	\caption{Uncertainty estimation for models on M-ARC. Agreement-based sample consistency was used to calculate Brier scores for top performing models. Models are grouped by model size, with larger models demonstrating relatively improved Brier scores compared to their smaller counterparts.}
	\label{fig:fig6}
\end{figure}

\section{Discussion}
Considering that the progression of AI development has continually drawn on insights from human cognition\cite{hassabis2017neuroscience, zador2023catalyzing, kumar2024shared} and that LLM training is reliant on extensive human-generated data, it is anticipated that LLMs may exhibit inductive biases that bear functional resemblance to cognitive biases in humans\cite{echterhoff2024cognitive, liu2024exploring, naeini2023large}. Characterizing such biases is essential to assessing their trustworthiness in clinical contexts. Here, we demonstrate that LLMs are vulnerable to the \textit{Einstellung} effect in medical QA tasks, where their inflexible adherence to matching learned statistical patterns impedes effective adaptation to medical scenarios that deviate from traditional medical texts and QA.

The disparity between LLM performance on M-ARC and conventional medical QA aligns with studies suggesting that benchmark successes may stem from overfitting to statistical patterns in training data rather than reflecting emergent reasoning abilities \citep{mccoy2023embers, goetz2024generalization, moskvichev2023conceptarc, dong2024generalization, li2024llms}. This interpretation is reinforced by studies that have shown LLMs' limited robustness and increased hallucination rates in low-probability contexts, where reliance on surface-level statistical correlations proves insufficient\citep{li2024llms, mccoy2023embers, yu2024hallucidoctor}. McCoy et al. hypothesized that poorer performance in these low-probability situations stem from biases inherent in the LLM training paradigm, which favors probabilistic strategies in autoregressive next-token prediction over development of robust deductive or abductive reasoning capabilities\citep{mccoy2023embers}. Consequently, LLMs may be biased toward surface-level correlations reinforced during training, which perform well on in-distribution data but hinder the development of generalizable reasoning strategies\citep{zhang2022paradox}, leaving them vulnerable to the \textit{Einstellung} effect\citep{naeini2023large}. M-ARC targets this inductive bias by disrupting the predictability of familiar medical problems through incorporation of long-tail concepts which are difficult for LLMs to capture effectively\citep{kandpal2023large}. 

Our findings align with recent evidence that LLMs have limited generalization---the ability to effectively apply reasoning to novel, out-of-distribution scenarios\citep{stechly2024chain, chollet2024arc}. Deficiencies of LLM generalization have been identified in multiple straightforward tasks\citep{nezhurina2024alice, li2024llms, arkoudas2023gpt}, including simple mathematical reasoning\citep{mirzadeh2024gsm}, and in planning, even with current state-of-the-art models\citep{stechly2024chain}. Generalization in out-of-distribution contexts is essential in real-world clinical scenarios, which often demand that reasoning strategies used in familiar, predictable situations be countermanded to consider more optimal approaches; this cognitive flexibility is foundational for effective clinical reasoning \citep{durning2016functional}. LLM inflexibility in reasoning, as demonstrated on M-ARC, indicates a lack of human-like cognitive flexibility and may hinder their ability to generalize to novel or unpredictable scenarios, undermining their reliability in real-world clinical contexts. 

Compounding these shortcomings are recently demonstrated LLM deficiencies in metacognition---specifically, the inability to recognize their own limitations---and overconfidence\citep{griot2025large}. Lack of metacognition and common sense in LLMs can lead to adverse outcomes if they are overrelied upon in clinical contexts\citep{kim2024m}. Our findings suggest that LLM limitations in reasoning may be exacerbated in long-tail or out-of-distribution contexts. To mitigate these risks, the development of selective prediction strategies, as proposed by Goetz et al.\citep{goetz2024generalization}, may offer a pathway to AI deployment in clinical scenarios. In this strategy, LLMs could defer to human clinicians in long-tail or out-of-distribution scenarios, ensuring that critical decisions are supervised by experts in contexts where LLMs may be unreliable\citep{goetz2024generalization}.

We acknowledge several limitations in this study. Compared to prior medical QA benchmarks such as specialty board exams and the USMLE, M-ARC consists of a smaller set of 100 questions. This reduced number reflects the nontrivial aspect of crafting questions that test long-tail or out-of-distribution reasoning patterns, which are more novel than those found in conventional medical QA. Future work will aim to increase the size of the M-ARC dataset to improve its robustness. In addition, we acknowledge that M-ARC questions are unlikely to be encountered in the real world. Nonetheless, our aim is not to benchmark or predict human competence in real-world clinical reasoning---an aim already addressed by existing assessments like the USMLE and board exams---but to probe failure modes in LLM reasoning in order to target areas where improvement is needed. Whereas conventional QA benchmarks are established at assessing human clinical reasoning, their reliability at evaluating LLM reasoning remains unclear\citep{griot2025large}. Lastly, we observe that human performance on M-ARC was limited, consistent with long-standing findings that humans may be susceptible to the \textit{Einstellung} effect. The average performance (~66\%)---comparable to typical accuracy on board examinations and in-training assessments\citep{mccrary2021systematic}---reflects the inherent variability in effort and reasoning abilities among subjects.

M-ARC reveals limitations in the medical reasoning of LLMs, challenging the notion that human-level performance on medical QA benchmarks suggests robust medical reasoning capabilities. The findings emphasize the need for the development of benchmarks that rigorously assess LLM generalization in medical reasoning, with assessment of reasoning flexibility serving as a potential approach. The observed shortcomings of LLMs in medical reasoning indicate the need for caution when utilizing LLMs in clinical contexts.

% \section{Supplementary Information}
% \label{sec:others}

% \subsection{Supplementary Tables}
% See awesome Table~\ref{tab:table}.

% The documentation for \verb+booktabs+ (`Publication quality tables in LaTeX') is available from:
% \begin{center}
% 	\url{https://www.ctan.org/pkg/booktabs}
% \end{center}

\bibliographystyle{unsrtnat}
\bibliography{references}

\end{document}